\documentclass[10pt,twocolumn,letterpaper]{article}

\usepackage{cvpr}
\usepackage{times}
\usepackage{graphics}
\usepackage{graphicx}
\usepackage{amsmath}
\usepackage{amssymb}
\usepackage{longtable,booktabs,array}
\usepackage{color}
\usepackage{subfigure}
\usepackage{multirow}
\usepackage{algorithm}
\usepackage{algorithmic}
\usepackage{epsfig}
\usepackage{verbatim}
\usepackage{paralist}
\usepackage{bm}

\usepackage[breaklinks=true,bookmarks=false]{hyperref}

\cvprfinalcopy 

\ifcvprfinal\fi
\begin{document}

\title{Recent Advances in Efficient Computation of \\
Deep Convolutional Neural Networks}
\author{
Jian Cheng$\thanks{The corresponding author.}$, Peisong Wang, Gang Li, Qinghao Hu, Hanqing Lu\\
National Laboratory of Pattern Recognition, Institute of Automation, Chinese Academy of Sciences \\
University of Chinese Academy of Sciences\\
{\tt\small jcheng@nlpr.ia.ac.cn}
}

\maketitle
\thispagestyle{empty}

\begin{abstract}
Deep neural networks have evolved remarkably over the past
few years and they are currently the fundamental tools of many intelligent systems.
At the same time, the computational complexity and resource consumption of these networks
also continue to increase. This will pose a significant challenge to the deployment
of such networks, especially in real-time applications or on resource-limited devices.
Thus, network acceleration has become a hot topic within the deep learning
community. As for hardware implementation of deep neural networks, a batch of accelerators 
based on FPGA/ASIC have been proposed in recent years. In this paper, we provide a 
comprehensive survey of recent advances in network acceleration, compression and accelerator 
design from both algorithm and hardware points of view.
Specifically, we provide a thorough analysis of each of the following topics: network pruning,
low-rank approximation, network quantization, teacher-student networks, compact network design
and hardware accelerators.
Finally, we will introduce and discuss a few possible future directions.
\end{abstract}


\section{Introduction}

In recent years, deep neural networks (DNNs) have achieved remarkable performance
across a wide range of applications, including but not limited to computer vision, natural language
processing, speech recognition, etc.
These breakthroughs are closely related to the increased
amount of training data and more powerful computing resources now available.
For example, one breakthrough in the natural image recognition field was achieved by AlexNet \cite{krizhevsky2012imagenet},
which was trained using multiple graphics processing units (GPUs) on about 1.2M images.
Since then, the performance of DNNs has continued to improve.
For many tasks, DNNs are reported to be able to outperform humans.
The problem, however, is that the computational complexity as well as the storage requirements
of these DNNs has also increased drastically as shown in Table \ref{tab:cnn}. 
Specifically, the widely used VGG-16 model \cite{simonyan2014very}
involves more than 500MB of storage and over 15B FLOPs to classify a single $224\times 224$ image.

Thanks to the recent crop of powerful GPUs and CPU clusters equipped with more abundant memory resources and computational units, these more powerful DNNs can be trained within a relatively reasonable time period.
However, when it is time for the inference phase, such a long execution time is impractical for real-time applications.
Recent years have witnessed great progress in embedded and mobile devices including unmanned drones, smart phones, intelligent glasses, etc.
The demand for deployment of DNN models on these devices has become more intense.
However, the resources of these devices, for example, the storage and computational units as
well as the battery power remain very limited, and this poses a real challenge in accelerating modern DNNs in low-cost settings.

Therefore, a critical problem currently is how to equip specific hardware with efficient deep networks without significantly lowering the performance. 
To deal with this issue, many great ideas and methods from the algorithm side have been investigated over the past few years. 
Some of these works focused on model compression while others focused on acceleration or lowering power consumption. 
As for the hardware side, a wide variety of FPGA/ASIC-based accelerators have been proposed for embedded and mobile applications.
In this paper, we present a comprehensive survey of several advanced approaches in network compression, acceleration and accelerator design. 
We will present the central ideas behind each approach and explore the similarities and differences between the different methods.
Finally, we will present some future directions in the field.

The rest of this paper is organized as follows. In Section \ref{sec:background}, we give some background on network acceleration and compression. From Section \ref{sec:pruning} to Section \ref{sec:net-design}, we systematically describe a series of hardware-efficient DNN algorithms, including network pruning, low-rank approximation, network quantization,
teacher-student networks and compact network design. In Section \ref{sec:hardware}, we introduce the design and implementation of hardware accelerators based on FPGA/ASIC technologies. In Section \ref{sec:future}, we discuss some future directions in the field, and Section \ref{sec:conclusion} concludes the paper.
 \begin{table*} [!ht]
\centering
\caption{The computation and parameters for state-of-art convolution neural networks}
\begin{tabular}{|c|c|c|c|c|c|c|}
\hline
\multirow{2}{*}{Method} & \multicolumn{3}{c|}{Parameters} & \multicolumn{3}{c|}{Computation}  \\ 
\cline{2-7} & Size(M) & Conv(\%)&Fc(\%)& FLOPS(G) & Conv(\%)&Fc(\%) \\
\hline AlexNet &61 &3.8 &96.2 &0.72 &91.9 &8.1 \\  
\hline VGG-S &103 &6.3 &93.7 &2.6 &96.3 &3.7 \\ 
\hline VGG16 &138 &10.6 &89.4 &15.5 &99.2 &0.8 \\ 
\hline NIN &7.6 &100 &0 &1.1 &100 &0\\ 
\hline GoogLeNet &6.9 &85.1 &14.9 &1.6 &99.9 &0.1 \\ 
\hline ResNet-18 &5.6 &100 &0&1.8 &100 &0 \\ 
\hline ResNet-50 &12.2 &100 &0 &3.8 &100 &0 \\ 
\hline ResNet-101 &21.2 &100 &0 &7.6 &100 &0 \\ 
\hline
\end{tabular}
\label{tab:cnn}
\end{table*}

\section{Background}
\label{sec:background}
  
Recently, deep convolutional neural networks (CNNs) have become quite popular due to their powerful 
representational capacity. With the huge success of CNNs, the demand for deployment of deep networks 
in real world applications has continued to increase. However, the large storage consumption and computational 
complexity remain two key problems for deployment of these networks. For the CNN training phase, 
the computational complexity is not a critical problem thanks to the high performance GPUs or CPU clouds.  
The large storage consumption also has less effect on the training phase because modern computers have very large disk and memory storage capacities. 
However, things are quite different for the inference phase in CNNs, especially with regard to embedded and mobile devices. 

The enormous computational complexity introduces two problems in the deployment of CNNs in real-world applications. 
One is that the CNN inference phase slows down as the computational complexity grows larger. 
This makes it difficult to deploy CNNs in real-time applications. 
The other problem is that the dense computation inherent to CNNs will consume substantial battery power, which is limited on mobile devices. 

The large number of parameters of CNNs consumes considerable storage and run-time memory, 
which are quite limited on embedded devices. In addition, it becomes more difficult to download new models online on mobile devices.  

To solve these problems, network compression and acceleration methods have been proposed. 
In general, the computational complexity of CNNs is dominated by the convolutional layers, 
while the number of parameters is mainly related to the fully connected layers as shown in Table \ref{tab:cnn}. 
Thus, most network acceleration methods focus on decreasing the computational complexity of the convolutional layers, 
while the network compression methods mainly try to compress the fully connected layers.

\section{Network Pruning}
\label{sec:pruning}
Pruning methods were proposed before deep learning became popular, and they have been widely studied in recent years \cite{lecun1989optimal,hassibi1993second,han2015deep,han2015learning}.
Based on the assumption that many parameters in deep networks are unimportant or unnecessary, 
pruning methods are used to remove the unimportant parameters. 
In this way, pruning methods can expand the sparsity of the parameters significantly. 
The high sparsity of the parameters after pruning introduces two benefits for deep neural networks. 
On the one hand, the sparse parameters after pruning require less disk storage since the parameters can be stored in the compressed sparse row format (CSR) or compressed sparse column (CSC) format. 
On the other hand, computations involving those pruned parameters are omitted; thus, the computational complexity of deep networks can be reduced. 
According to the granularity of the pruning, pruning methods can be categorized into five groups: 
fine-grained pruning, vector-level pruning, kernel-level pruning, group-level pruning and filter-level pruning. 
Figure.~\ref{pruning} shows the pruning methods with their different granularities. 
In the following subsections, we describe the different pruning methods in detail.
  \subsection{Fine-grained Pruning}
\begin{figure}
\centering
   \includegraphics[width=0.9\linewidth]{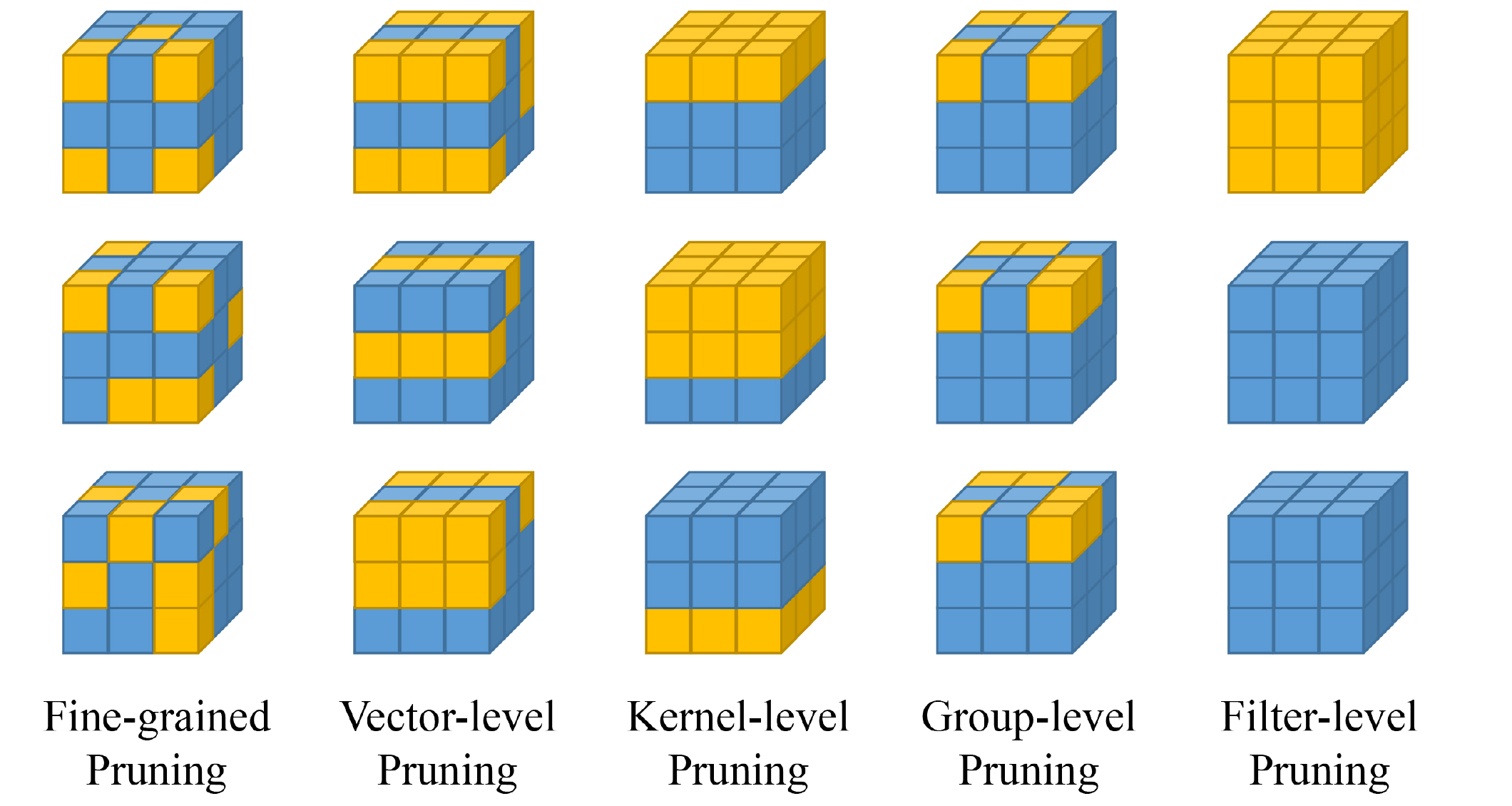}
   \caption{Different pruning methods for a convolutional layer which has 3 convolutional filters of size $3\times 3\times3$.}
\label{pruning}
\end{figure}

 Fine-grained pruning methods or vanilla pruning methods remove parameters in an unstructured way, 
 i.e., any unimportant parameters in the convolutional kernels can be pruned, as shown in Figure.~\ref{pruning}. 
 Since there are no extra constraints on the pruning patterns, the parameters can be pruned with a high sparsity. 
 Early works on pruning \cite{lecun1989optimal,hassibi1993second} used the approximate second-order derivativeses  
 of the loss function w.r.t. the parameters to determine the saliency of the parameters, 
 and then pruned those parameters with low saliency. 
 Yet, deep networks can ill afford to compute the second order derivativeses  due to the huge computational complexity. 
 Recently \cite{han2015deep} proposed a deep compression framework to compress deep neural networks in three steps: 
 pruning, quantization, and Huffman encoding. By using this method, AlexNet could be compressed by 35$\times$ without drops in accuracy. 
 After pruning, the pruned parameters in \cite{han2015deep} remain unchanged, incorrectly pruned parameters could cause accuracy drops. 
 To solve this problem, \cite{guo2016dynamic} proposed a dynamic network surgery framework,
 which consists of two operations: pruning and splicing. The pruning operation aims to prune those unimportant parameters while the splicing operation aims to recover the incorrectly pruned connections. Their method requires fewer training epochs and achieves a better compression ratio than \cite{han2015deep}.

 \subsection{Vector-level and Kernel-level Pruning}
 Vector-level pruning methods prune vectors in the convolutional kernels, and kernel-level pruning methods prune 2-D convolutional kernels in the filters. 
 Since most pruning methods focus on fine-grained pruning or filter-level pruning, there are few works on vector-level and kernel-level pruning.
 \cite{anwar2017structured} first explored the kernel-level pruning, and then proposed an intra-kernel strided pruning method,
 which prunes a sub-vector in a fixed stride.
  \cite{mao2017exploring} explored different granularity levels in pruning, and found that vector-level pruning takes up less storage 
  than fine-grained pruning because vector-level pruning requires fewer indices to indicate the pruned parameters. 
  Nevertheless, vector-level, kernel-level, and filter-level pruning techniques  are friendlier in hardware implementations since they are the more structured pruning methods.
\subsection{Group-level Pruning}
 \begin{figure}
\centering
   \includegraphics[width=0.9\linewidth]{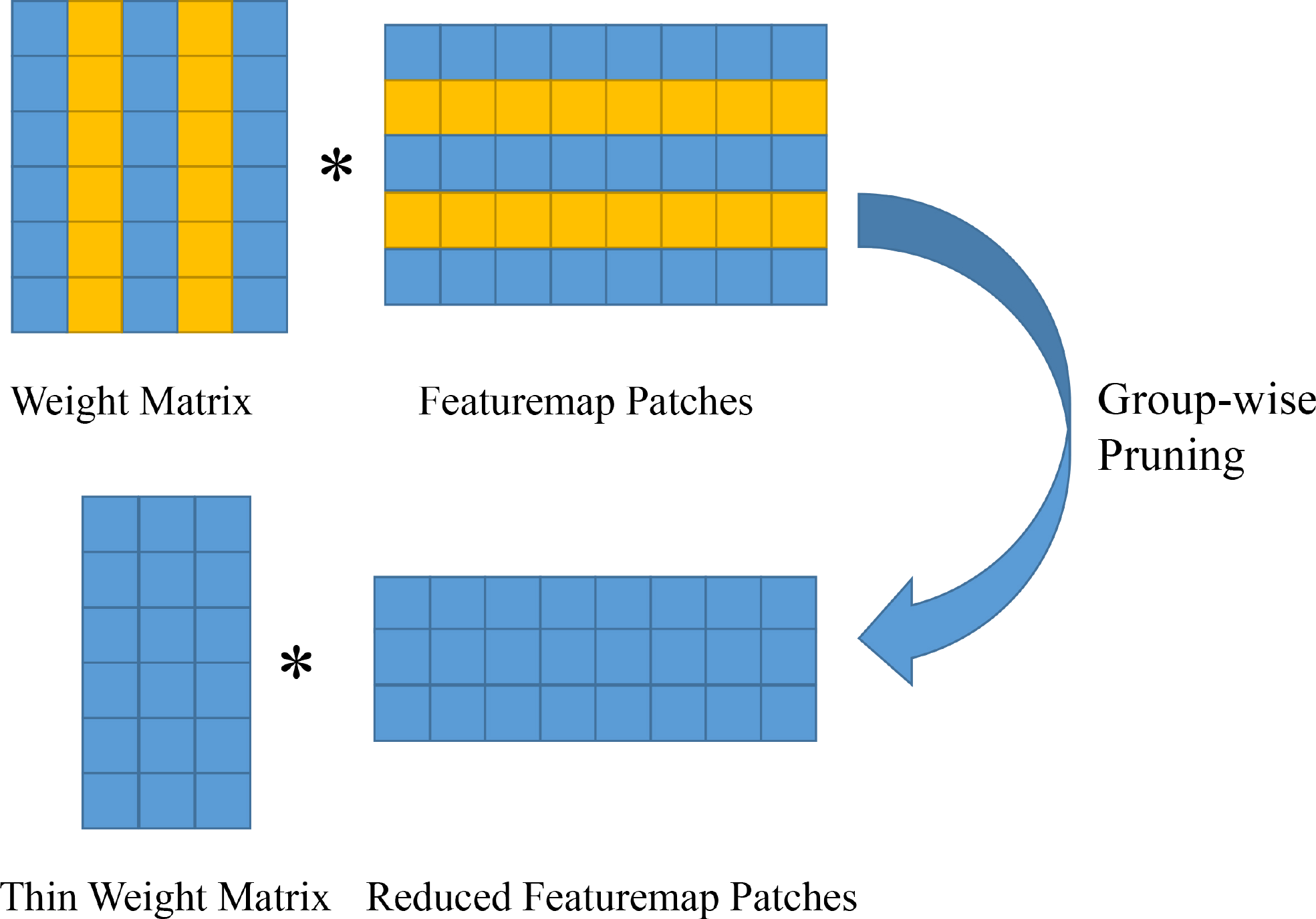}
   \caption{Group-level Pruning.}
\label{group-pruning}
\end{figure}
 Group-level pruning methods prune the parameters according to the same sparse pattern on the filters. As shown in Figure~\ref{group-pruning}, each filter has the same sparsity pattern, thus the convolutional filters can be represented as a thinned dense matrix. By using group-level pruning, convolutions can be implemented by thinned dense matrices multiplication. As a result, the Basic Linear Algebra Subprograms (BLAS) can be utilized to achieve a higher speed-up. \cite{lebedev2016fast} proposed the group-wise brain damage approach, which prunes the weight matrix in a group-wise fashion. By using group-sparsity regularization, deep networks can be trained easily with group-sparsified parameters. 
 Since group-level pruning can utilize the BLAS library, the practical speed-up is almost linear at the sparsity
level. By using this method, they achieved a 3.2$\times$ speed-up for all convolutional layers in AlexNet. 
Concurrent with \cite{lebedev2016fast}, \cite{wen2016learning} proposed using the group Lasso to prune groups of parameters. 
In contrast, \cite{wen2016learning} explored different levels of structured sparsity in terms of filters, channels, filter shapes, and depth. Their method can be regarded as a more general group-regularized pruning method. For AlexNet's convolutional layers, their method achieves about 5.1$\times$ and
$3.1\times$ speed-ups  on a CPU and GPU respectively.
 \subsection{Filter-level Pruning}
Filter-level pruning methods prune the convolutional filters or channels which make the deep networks thinner. 
After the filter pruning for one layer, the number of input channels of the next layer is also reduced.
Thus, filter-level pruning is more efficient for accelerating deep networks.
 \cite{luo2017thinet} proposed a filter-level pruning method named ThiNet. 
 They used the next layer's feature map to guide the filter pruning in the current layer. 
 By minimizing the feature map's reconstruction errors, they select the channels in a greedy way. 
 Similar to \cite{luo2017thinet}, \cite{he2017channel} proposed an iterative two-step algorithm to prune filters by minimizing the feature map errors. Specifically, they introduced a selection weight $\beta_i$ for each filter $\bm{W_i}$, then added sparse constraints on $\beta_i$. 
 Then the channel selection problem can be casted into a LASSO regression problem. 
 To minimize the feature map errors, they iteratively updated $\beta$ and $\bm{W}$. 
 Moreover, their method achieved a $5\times$ speed-up on VGG-16 network with little drop in accuracy. 
 Instead of using additional selection weight $\beta$,  \cite{liu2017learning} proposed to leverage the scaling factor of the batch normalization layer for
 to evaluate the importance of the filters. By pruning the channels with near-zero scaling factors, they can prune filters without introducing overhead into the networks.

\section{Low-rank Approximation}
\label{sec:low-rank}
The convolutional kernel of a convolutional layer $\bm{W}\in{R^{w\times h\times c\times n}}$ is a 4-D tensor.
These four dimensions correspond to the kernel width, kernel height and the number of input and output channels respectively.
Note that by merging some of the dimensions, the 4-D tensor
can be transformed into a $t$-D ($t=1,\cdots 4$) tensor.
The motivation behind low-rank decomposition is to find an approximate tensor $\hat{\bm{W}}$
that is close to $\bm{W}$ but facilitates more efficient computation.
Many low-rank based methods have been proposed by the community; two key differences are in
how to rearrange the four dimensions, and on which dimension the low-rank constraint is imposed.
Here we roughly divide the low-rank based methods into three categories, according to how many
components the filters are decomposed into: two-component decomposition, three-component
decomposition and four-component decomposition.

\subsection{Two-component Decomposition}
For two-component decomposition, the weight tensor is divided into two parts and the convolutional
layer is replaced by two successive layers.
\cite{jaderberg2014speeding} decomposed the spatial dimension $w*h$ into $w*1$ and $1*h$ filters.
They achieved a 4.5$\times$ speedup for a CNN trained on a text character recognition dataset, with a 1\%
accuracy drop.

SVD is a popular low-rank matrix decomposition method. By merging the
dimensions $w$, $h$ and $c$,
the kernel becomes a 2-D matrix of size $(w*h*c)\times n$, on which the SVD decomposition method can be conducted.
In \cite{denil2013predicting}, the authors utilized SVD to reduce the network redundancy.
SVD decomposition was also investigated in \cite{zhang2015accelerating}, in which the filters were
replaced by two filter banks: one consisting of $d$ filters of shape $w\times h \times c$ and the other
composed of $n$ filters of shape $1\times 1 \times d$.
Here, $d$ represents the rank of the decomposition, i.e., the $n$ filters are linear combinations of the first $d$
filters. They also proposed
the non-linear response reconstruction method based on the low-rank decomposition.
On the challenging VGG-16 model for the ImageNet classification task, this two-component SVD
decomposition method achieved a 3$\times$ theoretical speedup at a cost of
about 1.66\% increased top-5 error.

Similarly, another SVD decomposition method can be used by exploring the low-rank property
along the input channel dimension $c$. In this way, we reshape the weight tensor into a matrix
of size $c\times(w*h*n)$. By selecting the rank to $d$, the convolution can be decomposed
first by a $1\times 1 \times c \times d$ convolution and then by a $w\times h \times d \times n$ convolution.
These two decomposition are symmetric.

\subsection{Three-component Decomposition}
Based on the analysis of two-component decomposition methods,
one straightforward three-component decomposition method can be obtained by two successive
two-component decompositions.
Note that in the SVD decomposition, two weight tensors are introduced. The first is a
$w\times h \times c \times d$ tensor and the other is a $d \times n$ tensor (matrix).
The first convolution is also very time consuming due to the large size of the first tensor.
We can also conduct a two-component decomposition on the first weight tensor after
the SVD decomposition, which turns into a three-component decomposition method.
This strategy was studied by \cite{zhang2015accelerating}, whereby after the SVD decomposition,
they utilized the decomposition method proposed by \cite{jaderberg2014speeding} for the
first decomposed tensor. Thus, the final three components were convolutions with a spatial size
of $w\times 1$, $1 \times h$, and $1 \times 1$, respectively.
By utilizing this three-component decomposition, only a 0.3\% increased top-5 error was produced
in \cite{zhang2015accelerating} for a $4\times$ theoretical speedup.

If we use the SVD decomposition along the input channel dimension for the first tensor
after the two-component decomposition, we
can get the Tucker decomposition format as proposed by \cite{kim2015compression}.
These three components are convolutions of a spatial size $1\times 1$, $w\times h$ and another
$1\times 1$ convolution. Note that instead of using the two-step SVD decomposition, \cite{kim2015compression}
utilized the Tucker decomposition method directly to obtain these three components.
Their method achieved a 4.93$\times$ theoretical speedup at a cost of 0.5\% increased top-5 accuracy.

To further reduce complexity, \cite{wang2016accelerating} proposed a Block-Term Decomposition (BTD)
method based on low-rank and group sparse decomposition. Note that in the Tucker decomposition,
the second component corresponding to the $w\times h$ convolution also requires a large number of computations.
Because the second tensor is already low rank along both the input and output channel dimensions, the
decomposition methods discussed above cannot be used any longer. \cite{wang2016accelerating} proposed
to approximate the original weight tensor by the sum of some smaller subtensors, each of which is
in the Tucker decomposition format. By rearranging these subtensors, the BTD can be seen as a
Tucker decomposition where the second decomposed tensor is a block diagonal tensor.
By using this decomposition, they achieved a 7.4\% actual speedup for the VGG-16 model, at a cost
of a 1.3\% increased in the top-5 error. Their method also achieved high speedup for object detection and image retrieval
tasks as reported in \cite{wang2017deepsearch}.

\subsection{Four-component Decomposition}
By exploring the low-rank property along the input/output channel dimension as well as the spatial
dimension, a four-component decomposition can be obtained. This is corresponds to the
CP-decomposition acceleration method proposed in \cite{lebedev2014speeding}.
In this way, the four components are convolutions of size $1\times 1$, $w\times 1$, $1\times h$
and $1\times 1$. The CP-decomposition can achieve a very high speedup ratio, however,
due to the approximate error, only the second layer of AlexNet was processed in  \cite{lebedev2014speeding}.
They achieved a 4.5$\times$ speedup for the second layer of AlexNet at a cost of about a 1\% accuracy drop.

\begin{table*} [!ht] \small
\centering \caption{Comparison of fixed-point quantization methods according to which part is quantized and whether the training and testing
stages can be accelerated.}
\label{tab:fpq}
\begin{tabular}{|c|c|c|c|c|c|}
\hline
\multirow{2}{*}{Method} & \multicolumn{3}{c|}{Quantization} & \multicolumn{2}{c|}{Acceleration}  \\ 
\cline{2-6}                     &   Weight       &   Activation  & Gradient  & Training  & Testing  \\
\hline BinaryConnect \cite{courbariaux2015binaryconnect}    &   Binary        &   Full           &  Full         &  No            & Yes      \\  
\hline BWN \cite{rastegari2016xnor}                   &   Binary        &   Full           &  Full         &  No           &  Yes      \\  
\hline BWNH \cite{hu2018from}              &   Binary        &   Full           &  Full         &  No            &  Yes      \\  
\hline TWN \cite{li2016ternary}                   &   Binary        &   Full           &  Full         &  No            &  Yes      \\  
\hline FFN \cite{Wang_2017_CVPR}                     &   Ternary     &   Full           &  Full         &  No            &  Yes      \\  
\hline INQ \cite{zhou2017incremental}                     &   Ternary-5bit &   Full           &  Full         &  No            &  Yes      \\  
\hline BNN \cite{rastegari2016xnor}                   &   Binary       &   Binary        &  Full         &  No           &  Yes \\ 
\hline XNOR \cite{rastegari2016xnor}                 &   Binary       &   Binary        &   Full        & No            &   Yes  \\ 
\hline HWGQ \cite{Cai_2017_CVPR}          &   Binary       &   2bit           &   Full        & No            &   Yes  \\ 
\hline DoReFa-Net \cite{zhou2016dorefa}       &  Binary        & 1-4bit         &   6bit, 8bit, Full & Yes        &  Yes \\
\hline
\end{tabular}
\end{table*}

\section{Network Quantization}
\label{sec:quantization}
Quantization is an approach for many compression and acceleration applications. It has wide applications
in image compression, information retrieval, etc. Many quantization methods have also been investigated
for network acceleration and compression. We can categorize these methods into two main groups: (1) scalar and
vector quantization, which may need a codebook for quantization, and (2) fixed-point quantization.

\subsection{Scalar and Vector Quantization}

Scalar and vector quantization techniques have a long history, and they were originally used for data compression.
By using scalar or vector quantization, the original data can be represented by a codebook and a set of quantization codes. The codebook contains a set of quantization centers, and the quantization codes are used to indicate the assignment of the quantization centers. In general, the number of quantization centers is far smaller than the amount of original data. In addition,  quantization codes can be encoded through a lossless encoding method, e.g., Huffman coding, or just represented as low-bit fixed points. Thus, scalar or vector quantization can achieve a high compression ratio. \cite{gong2014compressing} explored scalar and vector quantization techniques for compressing deep networks. For scalar quantization, they used the well-known $K$-means algorithm to compress the parameters. In addition, the product quantization algorithm (PQ) \cite{jegou2011product}, a special case of vector quantization, was leveraged to compress the fully connected layers. By partitioning the feature space into several disjoint subspaces and then conducting $K$-means in each subspace, the PQ algorithm can compress the fully connected layers with little loss. As \cite{gong2014compressing} only compressed the fully connected layers, in \cite{wu2016quantized} and \cite{wu2017quantized}, the authors proposed to utilize the PQ algorithm to simultaneously accelerate and compress  convolutional neural networks. They proposed to quantize the convolutional filters layer by layer by minimizing the feature map's reconstruction loss. During the inference phase, a look-up table is built by pre-computing the inner product between feature map patches and codebooks, then the output feature map can be calculated by simply accessing the look-up table.
By using this method, they can achieve 4 $\sim$ 6$\times$ speedup and 15 $\sim$ 20$\times$ compression ratio with little accuracy loss.

\subsection{Fixed-point Quantization}
Fixed-point quantization is an effective approach for lowering the resource consumption of a network.
Based on which part is quantized, two main categories can be classified, i.e., weight quantization and activation quantization.
There are some other works that try to also quantize gradients, which can result in acceleration at the network training stage.
Here, we mainly review weight quantization and activation quantization methods, which accelerate the test-phase computation. 
Table \ref{tab:fpq} summarizes these methods according to which part is quantized and whether the training and testing
stages can be accelerated.

\subsubsection{Fixed-point Quantization of Weights}

Fixed-point weight quantization is a fairly mature topic in network acceleration and compression.
\cite{Hammerstrom2012A} proposed a VLSI architecture for network acceleration using
8-bit input and output, and 16-bit internal representation. In \cite{Holi1993Finite},
the authors provided a theoretical analysis of error caused by low-bit quantization to determine
the bitwidth for a multilayer perceptron. They showed that 8$\sim$16 bit quantization was sufficient for
training small neural networks. These early works mainly focused on simple multilayer perceptrons.
A more recent work \cite{Chen2014DaDianNao} showed that it is necessary to use 32-bit
fixed-point for the convergence of  a convolutional neural network trained on MNIST dataset.
By using stochastic rounding, the work by \cite{gupta2015deep} found that it is sufficient to use
16-bit fixed-point numbers to train a convolutional neural network on MNIST.
In addition, 8-bit fixed-point quantization was also investigated in \cite{dettmers20158} to speed up
the convergence of deep networks in parallel training.
Logarithmic data representation was also investigated in \cite{miyashita2016convolutional}.

Recently, much lower bit quantization or even binary and ternary quantization methods have been investigated.
Expectation Backpropagation (EBP) was introduced in \cite{cheng2015training}, which utilized the
variational Bayes method to binarize the network.
The BinaryConnect method proposed in \cite{courbariaux2015binaryconnect}
constrained all weights to be either +1 or -1.
By training from scratch, the BinaryConnect
can even outperform the floating-point counterpart on the CIFAR-10 \cite{krizhevsky2009learning}
image classification dataset.
Using binary quantization, the network can be
compressed about 32 times compared to 32-bit floating-point networks. Most of the floating-point
multiplication can also be eliminated \cite{lin2015neural}.
In \cite{rastegari2016xnor}, the authors proposed the Binary Weight Network (BWN),
which was among the earliest works that achieved good results on the large 
ImageNet \cite{russakovsky2015imagenet} dataset.
Loss-aware binarization was proposed in (\cite{hou2016loss}), which can directly minimize the classification loss with respect to the binarized weights.
In the work of \cite{hu2018from}, the authors proposed a novel approach
called BWNH to train Binary Weight Networks via Hashing, which outperformed other weight binarization methods by a large margin.
Ternary quantization was also utilized in \cite{hwang2014fixed}.
In \cite{li2016ternary}, the authors proposed the Ternary Weight Network (TWN), which was similar to BWN, but constrained all
weights to be ternary values among \{-1, 0, +1\}.
The TWN outperformed BWN by a large margin on deep models like ResNet.
Trained Ternary Quantization proposed in \cite{zhu2016trained} learned both ternary values
and ternary assignments at the same time using back-propagation. They achieved
comparable results on the AlexNet model.
Different from previous quantization methods, the Incremental Network Quantization (INQ) method
proposed in \cite{zhou2017incremental} gradually turned all weights into a logarithmic format
in a multi-step manner.
This incremental quantization strategy can lower the quantization error during each stage, and thus
can make the quantization problem much easier.
All these low-bit quantization methods discussed above directly quantize the full-precision
weight into a fixed-point format. In \cite{Wang_2017_CVPR}, the authors proposed a
very different quantization strategy. In stead of direct quantization, they proposed using
a fixed-point factorized network (FFN) to quantize all weights into ternary values.
This fixed-point decomposition method can significantly lower the quantization error.
The FFN method achieved comparable results on commonly used deep models such as
AlexNet, VGG-16 and ResNet.

\subsubsection{Fixed-point Quantization of Activations}
Given only weight quantization, there is also a need for the time-consuming floating-point
operations. If the activations were also quantized into fixed-point values, the network
can be efficiently executed by only fixed-point operations.
Many activation quantization methods were also proposed by the deep learning community.
The bitwise neural network was proposed in \cite{kim2016bitwise}.
Binarized Neural Networks (BNN) were among the first works that quantized both weights
and activations into either -1 or +1. BNN achieved a comparable accuracy with the full-precision
baseline on the CIFAR-10 dataset.
To extend the BNN for the ImageNet classification task, the authors in \cite{tang2017train}
improved the training strategies of the BNN. Much higher accuracy was reported using these strategies.
Based on the BWN, the authors in \cite{rastegari2016xnor} further quantize all activations
into binary values, making the network into a XNOR-Net. Compared with BNN, the XNOR-Net
can achieve much higher accuracy on the ImageNet dataset.
To further understand the effect of bit-width on the training of deep neural networks, the DoReFa-Net
was proposed in \cite{zhou2016dorefa}.
It investigated the effect of different bit-widths for weights and activations as well as gradients.
By making use of batch normalization, the work by \cite{Cai_2017_CVPR} presented
the Half-wave Gaussian Quantization (HWGQ) method to quantize both weights and activations.
A high performance was achieved on commonly
used CNN models using the HWGQ methond, with 2-bit activations and binary weights.

\section{Teacher-student Network}
\label{sec:teacher-student}
The teacher-student network is different from the network compression or acceleration methods since it trains a student network using a teacher network and the student network can be designed with a different network architecture. Generally speaking, a teacher network is a large neural network or the ensemble of neural networks while a student network is a compact and efficient neural network. By utilizing the dark knowledge transferred from the teacher network, the student network can achieve higher accuracy than training merely through the class labels.  \cite{hinton2015distilling} proposed the knowledge distillation (KD) method which trains a student network by the softmax layer's output of the teacher network. Following this line of thinking, \cite{romero2014fitnets} proposed the FitNets to train a deeper and thinner student network. Since the depth of neural networks is more important than the width of them, a deeper student network would have higher accuracy. Besides, they utilized both intermediate layers' featuremaps and soft outputs of the teacher network to train the student network. Rather than mimicking the intermediate layers' feature maps, \cite{zagoruyko2016paying} proposed to train a student network by imitating the attention maps of a teacher network. Their experiments showed that the attention maps are more important than the layers' activations and their method can achieve higher accuracy than FitNets.

\section{Compact Network Design}
\label{sec:net-design}

The objective of network acceleration and compression is to optimize the execution and
storage framework for a given deep neural network. One property is that the network architecture
is not changed. Another parallel line of inquiry for network acceleration and compression
is to design more efficient but low-cost network architecture itself.

In \cite{lin2013network}, the authors proposed Network-In-Network architecture, where
a $1\times 1$ convolution was utilized to increase the network capacity while keeping the
overall computational complexity small. To reduce the storage requirement of the CNN models,
they also proposed to remove the fully connected layer and make use of a global average pooling.
These strategies are also used by many state-of-the-art CNN models like GoogLeNet \cite{Szegedy2015Going} and
ResNet \cite{he2015deep}.

Branching (multiple group convolution) is another commonly used strategy for lowering
network complexity, which was explored in the work of GoogLeNet \cite{Szegedy2015Going}.
By largely making use of $1\times 1$ convolution and the branching strategy, the SqueezeNet proposed
in \cite{iandola2016squeezenet} achieved about 50$\times$ compression over AlexNet, with
comparable accuracy. By branching, the work of ResNeXt work of \cite{Xie_2017_CVPR} can achieve much higher
accuracy than the ResNet \cite{he2015deep} at the same computational budget.
The depth-wise convolution proposed in MobileNet \cite{howard2017mobilenets}
takes the branching strategy to the extreme, i.e., the number of branches equals the number
of input/output channels. The resulting MobileNet can be 32$\times$ smaller
and 27$\times$ faster than the VGG-16 model, with comparable image classification accuracy on ImageNet.
When using depth-wise convolution and $1\times 1$ convolution as in MobileNet,
most of the computation and parameters reside in the $1\times 1$ convolutions.
One strategy to further lower the complexity of the $1\times 1$ convolution is to use multiple groups.
The ShuffleNet proposed in \cite{zhang2017shufflenet} introduced the channel shuffle operation to
increase the information change within the multiple groups, which can prominently increase
the representational power of the networks. Their method achieved about 13$\times$ actual speedup
over AlexNet with comparable accuracy.

\section{Hardware Accelerator}
\label{sec:hardware}

\subsection{Background}
Deep neural networks provide impressive performance for various tasks while suffering from degrees of  computational complexity. Traditionally, algorithms based on deep neural networks should be executed on general purpose platforms such as CPUs and GPUs, but this works at the expense of unexpected power consumption and oversized resource utilization for both computing and storage. In recent years, there are an increasing number of applications that are based on embedded systems, including autonomous vehicles, unmanned drones, security cameras, etc. Considering the demands for high performance, light weight and low power consumption on these devices, CPU/GPU-based solutions are no longer suitable. In this scenario, FPGA/ASIC-based hardware accelerators are gaining popularity as efficient alternatives.

\subsection{General Architecture}
\begin{figure}
  \centering
  \includegraphics[width=\columnwidth]{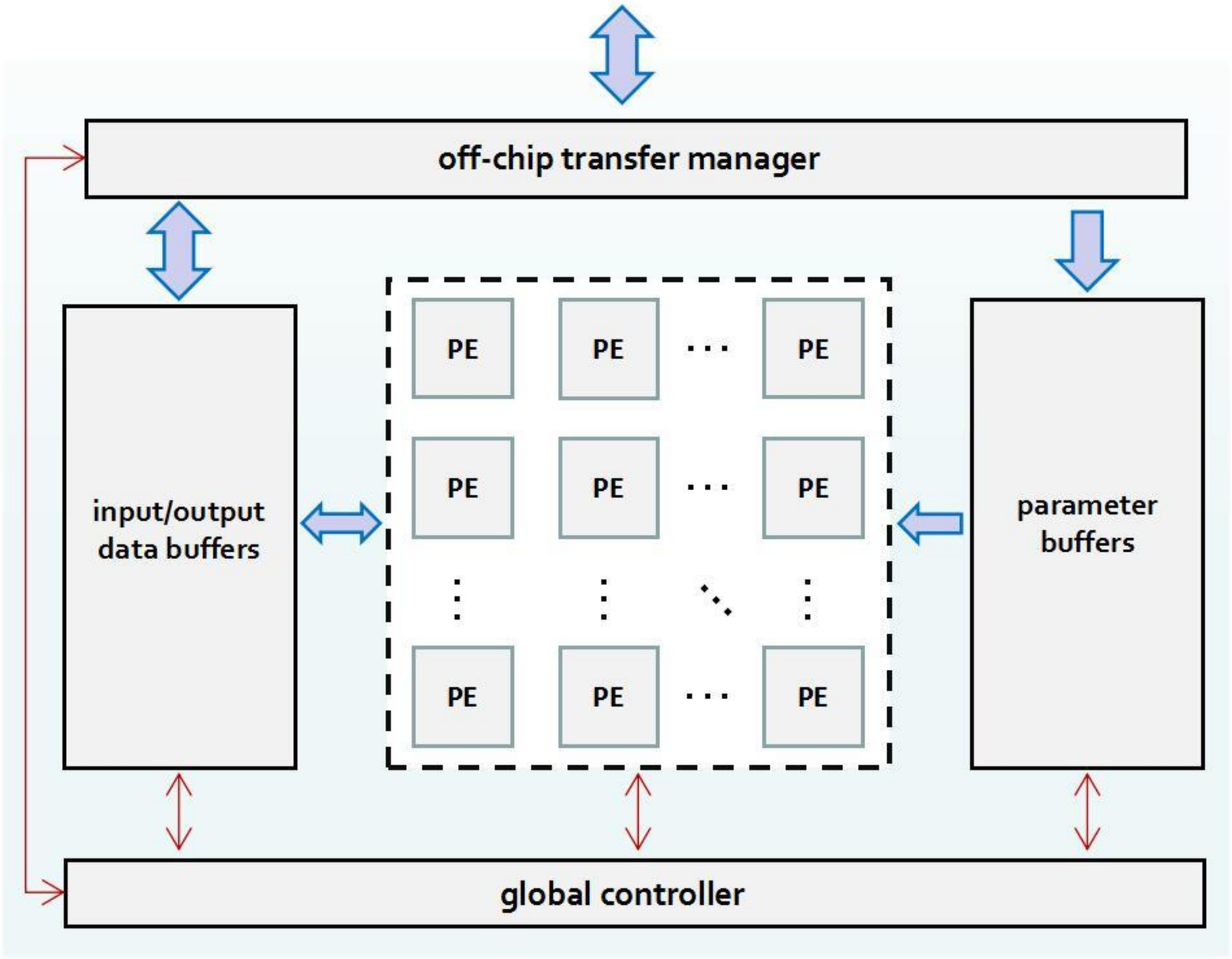}\\
  \caption{General architecture of an accelerator on dedicated hardware.}\label{arch}
\end{figure}

The deployment of a DNN on a real-world application consists of two phases: training and inference. Network training is known to be expensive in terms of speed and memory, thus it is usually carried out on GPUs off-line. During the inference phase, the pre-trained network parameters can be loaded either from the cloud or from dedicated off-chip memory. Most recently, hardware accelerators for training have received widespread attention \cite{Ko2017Design, Yang2017TIME, Venkataramani:2017:SSC:3140659.3080244}, but in this section we mainly focus on the inference phase in embedded settings.

Typically, an accelerator is composed of five parts: data buffers, parameter buffers, processing elements, global controller and off-chip transfer manager, as shown in
Figure.~\ref{arch}.
The data buffers are used to caching input images, intermediate data and output predictions, while the weight buffers are mainly used to cache convolutional filters. Processing elements are a collection of basic computing units that execute multiply-adds, non-linearity and any other functions such as normalization, quantization, etc. The global controller is used to orchestrate the computing flow on-chip, while off-chip transfers of data and instructions are conducted through a manager. This basic architecture can be found in existing accelerators designed for both specific and general tasks.

Heterogeneous computing is widely adopted in hardware acceleration. For computing-intensive operations such as multiply-adds, it is efficient to fit them on hardware for high throughout, otherwise, data pre-processing, softmax and any other graphic operations can be placed on the CPU/GPU for low latency processing.

\subsection{Processing Elements}

Among all of the accelerators, the biggest differences exist in the processing elements as they are designed for the majority of computing tasks in deep networks, such as massive multiply-add operations, normalization (batch normalization or local response normalization), and non-linearities (ReLU, sigmoid and tanh). Typically, the computing engine of an accelerator is composed of many small basic processing elements, as shown in
Figure.~\ref{arch},
and this architecture is mainly designed for fully investing in data reuse and parallelism. However, there are many accelerators that operate with only one processing element in consideration of lower data movement and resource conservation \cite{Zhang:2015:OFA:2684746.2689060, Ma2017End}.
\subsection{Optimizing for High Throughput}

Since the majority of the computations in a network are matrix-matrix/matrix-vector multiplication, it is critical to deal with the massive nested loops to achieve high throughput.
Loop optimization is one of the most frequently adopted techniques in accelerator design \cite{Zhang:2015:OFA:2684746.2689060, Ma:2017:OLO:3020078.3021736, Suda:2016:TOF:2847263.2847276, Alwani2016FusedlayerCA, Xiao:2017:EHA:3061639.3062244, blockconvgli}, including loop tiling, loop unrolling, loop interchange, etc. Loop tiling is used to divide all of the data into multiple small blocks in order to alleviate the pressure of on-chip storage \cite{Ma:2017:OLO:3020078.3021736, Alwani2016FusedlayerCA, Qiu:2016:GDE:2847263.2847265}, while loop unrolling attempts to improve the parallelism of the computing engine for high speed \cite{Ma:2017:OLO:3020078.3021736, Qiu:2016:GDE:2847263.2847265}. Loop interchange determines the sequential computation order of the nested loops because different computation orders can result in significant differences in performance.
The well-known \emph{systolic array} can be seen as a combination of the loop optimization methods listed above, which leverage the nature of data locality and weight sharing in the network to achieve high throughput \cite{Jouppi:2017:IPA:3079856.3080246, Wei:2017:ASA:3061639.3062207}.

SIMD-based computation is another way for high throughput. \cite{DBLP:conf/date/NguyenKL17} presented a method for packing two low-bit multiplications into a single DSP block to double the computation, and \cite{Price201714} also proposed a SIMD-based architecture for speech recognition.

\subsection{Optimizing for Low Energy Consumption}

Existing works attempt to reduce the energy consumption of a hardware accelerator from both computing and I/O perspectives. \cite{Horowitz20141} systematically illustrated the energy cost in terms of arithmetic operations and memory accesses. He demonstrated that operations based on integers are much more cheaper than their float-point counterparts, and lower bit integers are better. Therefore, most existing accelerators adopt low-bit or even binary data representation \cite{Zhao:2017:ABC:3020078.3021741, Umuroglu:2017:FFF:3020078.3021744, Nurvitadhi:2017:FBG:3020078.3021740} to preserve energy efficiency. Most recently, logarithmic computation that transfers multiplications into bit shift operations has also shown its promise in energy savings \cite{Lee2017LogNetEN, ShiftCNN, DBLP:journals/corr/TannHBR17}.

Sparsity is gaining an increased popularity in accelerator design based on the observation that a great number of arithmetic operations can be discarded to obtain energy efficiency. \cite{Han2016EIE}, \cite{Han2017ESE} and \cite{Parashar2017SCNN} designed architectures for image or speech recognition based on network pruning, while \cite{Albericio2016Cnvlutin} and \cite{Zhang2016Cambricon} proposed to eliminate ineffectual operations based on the inherent sparsity in networks.

Off-chip data transfers happen inordinately in hardware accelerators due to the fact that both network parameters and intermediate data are too large to fit on chip. \cite{Horowitz20141} suggested that power consumption caused by DRAM access is several orders of magnitude of the SRAM, and therefore reducing off-chip transfers is a critical issue. \cite{Shen2017Escher} designed a flexible data buffing scheme to reduce bandwidth requirements, and \cite{Alwani2016FusedlayerCA} and \cite{Xiao:2017:EHA:3061639.3062244} proposed a fusion-based method to reduce off-chip traffic. Most recently, \cite{blockconvgli} presented a block-based convolution that can completely avoid off-chip transfers of intermediate data in VGG-16 with high throughput.

Many other approaches have been proposed to reduce power consumption. \cite{Zhang:2016:ECI:2934583.2934644} used a pipelined FPGA cluster to realize acceleration, \cite{Chen2017Eyeriss} presented an energy-efficient row stationary scheme to reduce data movements, and \cite{Zhu2016LRADNN} attempted to reduce power consumption via low-rank approximation.

\subsection{Design Automation}

Recently, design automation frameworks that automatically map deep neural networks onto hardware are receiving wider attention. \cite{Wang2016DeepBurning}, \cite{Sharma2016From}, \cite{Venieris2016fpgaConvNet} and \cite{Wei:2017:ASA:3061639.3062207} proposed frameworks that automatically generate synthesizable accelerator for a given network. \cite{Ma2017An} presented an RTL compiler for FPGA implementation of diverse networks. \cite{Liu:2016:CIS:3007787.3001179} proposed an instruction set for hardware implementation, while \cite{Zhang2016Caffeine} proposed a uniformed convolutional matrix multiplication representation for CNNs.

\subsection{Emerging Techniques}

In the past few years, there have been many new techniques from both the algorithm side and the circuit side that have been adopted to implement fast and energy-efficient accelerators. Stochastic computing representing continuous values through streams of random bits have been investigated for hardware acceleration of deep neural networks \cite{Ren2017SC, Sim2017A, Kim2016Dynamic}. On the hardware side, RRAM-based accelerators
\cite{Chen2017Accelerator, Xia2016Switched} and the usage of 3-D DRAM \cite{Kim2016Neurocube, Gao2017TETRIS} have received greater attention.

\section{Future Trends and Discussion}
\label{sec:future}
In this section, we discuss some possible future directions in this field.
\paragraph{Non-fine-tuning or Unsupervised Compression.}
Most of the existing methods, including network pruning, low-rank compression and quantization,
need labeled data to retrain the network for accuracy retention. The problems are twofold.
First, labeled data is sometimes unavailable, as in medical images. Another problem is that
retraining requires considerable human effort as well as professional knowledge. These two problems
raise the need for unsupervised compression or even fine-tuning-free compression methods.

\paragraph{Scalable (Self-adaptive) Compression.}
Current compression methods have many hyperparameters that need to be determined ahead of time.
For example, the sparsity of the network pruning, the rank of the decomposition-based methods or
the bitwidth of fixed-point quantization methods. The selection of these hyperparameters
is tedious work, which also requires professional experience. Thus, the investigation of methods
that do not rely on human-designed hyperparameters is a promising research topic.
One direction may be to use annealing methods, or reinforcement learning.

\paragraph{Network Acceleration for Object Detection.}
Most of the model acceleration methods are optimized for image classification, yet very little effort has been devoted to the acceleration of other computer vision tasks such object detection. It seems that model acceleration methods for image classification can be directly used for detection. However, the deep neural networks for object detection or image segmentation are more sensitive to model acceleration methods, i.e., using the same model acceleration methods for object detection would suffer from a greater number of accuracy drops than with image classification. One reason for this phenomenon may be that object detection requires more complex feature representation than image classification.
The design of model acceleration methods for object detection represents a challenge.

\paragraph{Hardware-software Co-design.}
To accelerate deep learning algorithms on dedicated hardware, a straightforward method is to pick up
a model and design a corresponding architecture. However, the gap between algorithm modeling and hardware implementation will make it difficult to put this into practice. Recent advances in deep learning algorithms and hardware accelerators demonstrate that it is highly desirable to design hardware-efficient algorithms
according to the low-level features of specific hardware platforms. This co-design methodology will be a trend in future work.

\section{Conclusion}
\label{sec:conclusion}
Deep neural networks provide impressive performance while suffering from huge computational complexity and high energy expenditure. In this paper, we provide a survey of recent advances in efficient processing of deep neural networks from both the algorithm and hardware points of view. In addition, we point out a few topics that deserve further investigation in the future.

{\small
\bibliographystyle{ieee}
\bibliography{mybibfile}
}

\end{document}